\documentclass[letterpaper, 10 pt, conference]{ieeeconf}  

\IEEEoverridecommandlockouts                              

\title{KM-ViPE: Online Tightly Coupled Vision-Language-Geometry Fusion for Open-Vocabulary Semantic SLAM}

\usepackage{tikz}
\def\checkmark{\tikz\fill[scale=0.4](0,.35) -- (.25,0) -- (1,.7) -- (.25,.15) -- cycle;} 

\usepackage{mathtools, nccmath}
\usepackage{amsmath}
\usepackage{amssymb}
\usepackage{algorithm}
\usepackage{algpseudocode}
\usepackage{bm}
\usepackage{graphicx}
\usepackage{booktabs}

\author{Zaid Nasser$^{1}$, Mikhail Iumanov$^{1}$, Tianhao Li$^{1}$ , Maxim Popov$^{1}$, \\Jaafar Mahmoud$^{1,2}$, Malik Mohrat$^{1,2}$, Ilya Obrubov$^{2}$, Ekaterina Derevyanka$^{2}$, Ivan Sosin$^{2}$,\\ Sergey Kolyubin$^{1}$
\thanks{$^{1}$ Biomechatronics and Energy-Efficient Robotics (BE2R) Lab, ITMO
University, Saint Petersburg, Russia
        }%
\thanks{$^{2}$ SBERRoboticsCenter, Moscow, Russia
        }%
}
\newcommand{\name}{KM-ViPE}
\usepackage{threeparttable}  
\usepackage{xcolor}
\usepackage{pifont}

\definecolor{traditional}{RGB}{0, 150, 150}  
\definecolor{semslam}{RGB}{0, 120, 150}  
\definecolor{semantic}{RGB}{230, 159, 0}     
\definecolor{neural}{RGB}{153, 51, 255}      
\definecolor{ours}{RGB}{0, 128, 0}           
\definecolor{darkcyan}{rgb}{0.0, 0.2, 0.13}
\usepackage{graphicx}
\usepackage{booktabs}
\begin{document}

\maketitle
\thispagestyle{empty}
\pagestyle{empty}


\maketitle
\begin{abstract}

We present KM-ViPE (\textbf{K}nowledge \textbf{M}apping \textbf{Vi}deo \textbf{P}ose \textbf{E}ngine), a real-time open-vocabulary SLAM framework for uncalibrated monocular cameras in dynamic environments. Unlike systems requiring depth sensors and offline calibration, KM-ViPE operates directly on raw RGB streams, making it ideal for ego-centric applications and harvesting internet-scale video data for training.
KM-ViPE tightly couples DINO visual features with geometric constraints through a high-level features based adaptive robust kernel that handles both moving objects and movable static objects (e.g., moving furniture in ego-centric views). The system performs simultaneous online localization and open-vocabulary semantic mapping by fusing geometric and deep visual features aligned with language embeddings.
Our results are competitive with state-of-the-art approaches, while existing solutions either operate offline, need depth data and/or odometry estimation, or lack dynamic scene robustness. KM-ViPE benefits from internet-scale training and uniquely combines online operation, uncalibrated monocular input, and robust handling of dynamic scenes, which makes it a good fit for autonomous robotics and AR/VR applications and advances practical spatial intelligence capabilities for embodied AI\footnotemark[3]\footnotetext[3]{ https://github.com/be2rlab/km-vipe}.

\end{abstract}    
\section{Introduction}
\label{sec:intro}
Conventional SLAM systems focus on reconstruction and localization using geometric primitives such as points, lines, and planes~\cite{ORB-SLAM3, rvwo, airslam}. While effective in many settings, these systems capture only geometric scene representations and also degrade in dynamic environments, where moving objects confound data association across frames. With recent developments of general-purpose robots, spatial context-aware navigation in diverse environments is crucial. Open-vocabulary semantic perception enables it by allowing for interpretation beyond fixed taxonomies \cite{semprob}, which improves reasoning behind interaction. Yet, integrating rich semantics into SLAM without sacrificing real-time performance under scene dynamics remains a challenge. 

In addition, training navigation policies solely from curated robot logs constrains coverage, diversity, and long-tail phenomena. In contrast, Internet-scale videos, especially ego-centric data, expose agents to vast scene layouts, object appearances, and human activities in both indoor and outdoor environments, yielding visual priors that can be transferred to embodied control \cite{ego4d, aria}. However, real world environments challenge navigation not only with moving agents (people, pets, vehicles) but also with objects that are usually classified as static, but can be moved like furniture, kitchen appliances and dishes, room decorations, etc. This can corrupt data associations, degrade pose estimation, and pollute maps if not filtered explicitly. Together, these trends motivate harvesting Internet-scale videos to pre-train navigation policies with broad, robust semantic priors, then coupling them with SLAM systems that detect both moving and moved objects. This combination promises better reconstruction and navigation in dynamic environments where autonomous agents must operate.

We introduce \name{}, the \textbf{K}nowledge \textbf{M}apping \textbf{Vi}deo \textbf{P}ose \textbf{E}ngine, an online SLAM framework that unifies a geometric backbone with robust deep visual features from foundation models~\cite{dinov2,dinov3}. \name{} produces a language-interactive, open-vocabulary semantic map while simultaneously delivering metrically consistent 3$\mathrm{D}$ reconstruction and camera trajectory estimation in dynamic, real-world, egocentric settings.

\begin{figure}[t]
    \centering
    \includegraphics[width=\linewidth]{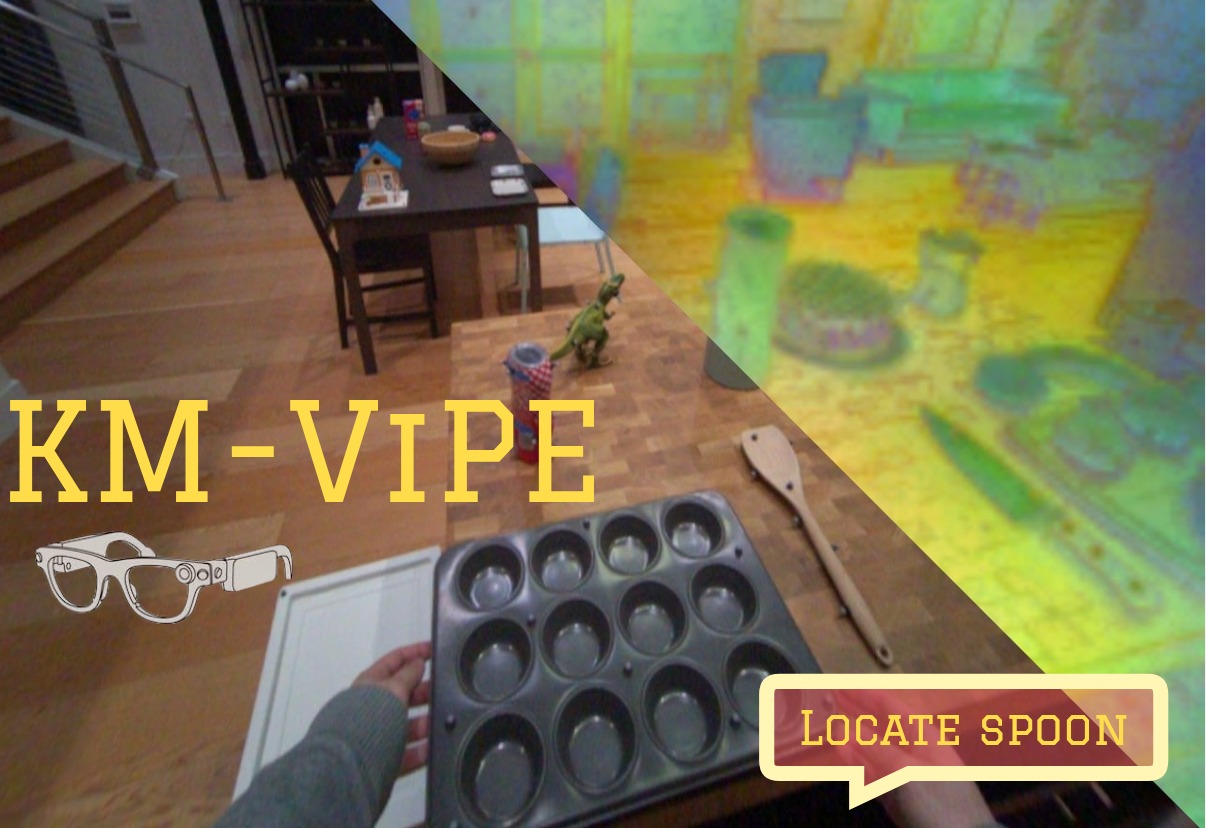}
    \caption{KM-ViPE: a real-time open-vocabulary SLAM framework for uncalibrated monocular RGB cameras in dynamic environments for Ego-Centric applications.}
    \label{fig:teaser}
\end{figure}
Following~\cite{vipe}, \name{} processes raw RGB streams from an uncalibrated monocular camera. We estimate camera intrinsics by sampling frames with GeoCalib~\cite{geocalib} and select keyframes based on optical flow motion criteria~\cite{droid}. For each keyframe, we extract dense DINO descriptors~\cite{dinov2,dinov3} and estimate metric depth~\cite{moge, unidepth, metric3dv2}. For efficient pixel-level representation, we use pyramid upsampling with PCA-encoded features. Dense embeddings tune an adaptive robust kernel~\cite{ark} that weights optical flow error terms, effectively handling moving objects under geometric constraints. Descriptors, optical flow, and depth are fused across keyframes, integrating visual and geometric cues. We construct a pose graph over recent keyframes based on temporal proximity and co-visibility. Camera poses and 3$\mathrm{D}$ points are optimized via Gauss–Newton minimization of a visual-geometric energy function. The resulting multi-modal point cloud supports open-vocabulary queries by decoding visual representations and projecting them into a joint vision-language space~\cite{talk2dino}.
The main contributions of this work are stated as follows:
\begin{enumerate}
\item A novel multi-modal fusion approach that integrates high-level visual representations and geometric constraints within a tightly-coupled framework. We combine robust pixel-level visual features from DINO \cite{dinov2, dinov3} with SLAM geometric information to create semantically-rich representations that maintain both metric accuracy and open-vocabulary accessibility.
\item An adaptive dynamic scene handling adjustment based on adaptive robust kernels and semantic error terms for map reconstruction. Our approach detects and filters moving objects—including semantically static entities that have been physically displaced (e.g., moved chairs, books, or cups) ensuring robust mapping performance in challenging dynamic environments specifically tailored for ego-centric environments.
\item A complete online open-vocabulary semantic SLAM system that unifies vision, language, and geometry to enable language-interactive spatial understanding from uncalibrated RGB input, advancing practical spatial intelligence for embodied agents by providing robust semantic priors suited for navigation and ego-centric robotic perception—even when learned from Internet-scale video data.
\end{enumerate}

\begin{table}[t]
  \centering
  \small
  \begin{threeparttable}

    \begin{tabular}{@{}lccccccc@{}}
      \toprule
      \textbf{Method} & \rotatebox{90}{Online} & \rotatebox{90}{Semantics} & \rotatebox{90}{Grounding} & \rotatebox{90}{Odometry} & \rotatebox{90}{Mapping} & \rotatebox{90}{Dynamic} & \rotatebox{90}{Calib-free}  \\
      \midrule
      
      \textcolor{traditional}{ORBS3}~\cite{ORB-SLAM3}       & \textcolor{olive}{\checkmark} & \textcolor{red}{\ding{55}}    & \textcolor{red}{\ding{55}}    & \textcolor{olive}{\checkmark} & \textcolor{olive}{\checkmark} & \textcolor{red}{\ding{55}} & \textcolor{red}{\ding{55}} \\
      \textcolor{traditional}{RVWO}~\cite{rvwo}       & \textcolor{olive}{\checkmark} & \textcolor{red}{\ding{55}}    & \textcolor{red}{\ding{55}}    & \textcolor{olive}{\checkmark} & \textcolor{olive}{\checkmark} & \textcolor{olive}{\checkmark} &\textcolor{red}{\ding{55}} \\
      \textcolor{traditional}{Kimera}~\cite{kimera}             & \textcolor{olive}{\checkmark} & \textcolor{red}{Closed}       & \textcolor{red}{\ding{55}}    & \textcolor{olive}{\checkmark} & \textcolor{olive}{\checkmark} & \textcolor{red}{\ding{55}} &\textcolor{red}{\ding{55}} \\
      \textcolor{traditional}{SamS}~\cite{SamSLAM}                & \textcolor{olive}{\checkmark} & \textcolor{blue}{Agnostic}     & \textcolor{red}{\ding{55}}    & \textcolor{olive}{\checkmark} & \textcolor{olive}{\checkmark} & \textcolor{olive}{\checkmark} &\textcolor{red}{\ding{55}}\\
      \midrule
      
      \textcolor{semantic}{BBQ}~\cite{bbq}                      & \textcolor{red}{\ding{55}}      & \textcolor{green}{Open}     & \textcolor{olive}{\checkmark} & \textcolor{red}{\ding{55}}      & \textcolor{olive}{\checkmark} & \textcolor{red}{\ding{55}}&\textcolor{red}{\ding{55}} \\
      \textcolor{semantic}{CG}~\cite{concept-graphs} & \textcolor{red}{\ding{55}}      & \textcolor{green}{Open}     & \textcolor{olive}{\checkmark} & \textcolor{red}{\ding{55}}      & \textcolor{olive}{\checkmark} & \textcolor{red}{\ding{55}} &\textcolor{red}{\ding{55}}\\
      \textcolor{semantic}{HOV-SG}~\cite{hovsg}        & \textcolor{red}{\ding{55}}      & \textcolor{green}{Open}     & \textcolor{olive}{\checkmark} & \textcolor{red}{\ding{55}}      & \textcolor{olive}{\checkmark} & \textcolor{red}{\ding{55}} &\textcolor{red}{\ding{55}}\\
      
      \textcolor{semantic}{OM3D}~\cite{openmask3d}          & \textcolor{red}{\ding{55}}      & \textcolor{green}{Open}     & \textcolor{olive}{\checkmark} & \textcolor{red}{\ding{55}}      & \textcolor{red}{\ding{55}}    & \textcolor{red}{\ding{55}} &\textcolor{red}{\ding{55}}\\
      \midrule

      \textcolor{neural}{CLIO}~\cite{clio}                    & \textcolor{olive}{\checkmark} & \textcolor{orange}{Task-driven} & \textcolor{olive}{\checkmark} & \textcolor{olive}{\checkmark} & \textcolor{olive}{\checkmark} & \textcolor{red}{\ding{55}} &\textcolor{red}{\ding{55}}\\
      \textcolor{neural}{OVO}~\cite{OVO-SLAM}         & \textcolor{olive}{\checkmark} & \textcolor{green}{Open}     & \textcolor{olive}{\checkmark} & \textcolor{olive}{\checkmark} & \textcolor{olive}{\checkmark} & \textcolor{red}{\ding{55}} &\textcolor{red}{\ding{55}}\\
      \midrule
      \textbf{\textcolor{purple}{Mast3R}~\cite{mast3r}}& 
      \textcolor{olive}{\checkmark} & \textcolor{red}{\ding{55}}      & \textcolor{red}{\ding{55}}  & \textcolor{olive}{\checkmark} & \textcolor{olive}{\checkmark} & \textcolor{red}{\ding{55}}  &\textcolor{olive}{\checkmark}\\
      \textbf{\textcolor{purple}{VGGT}~\cite{vggt}}& 
      \textcolor{olive}{\checkmark} & \textcolor{red}{\ding{55}}      & \textcolor{red}{\ding{55}}  & \textcolor{olive}{\checkmark} & \textcolor{olive}{\checkmark} & \textcolor{red}{\ding{55}}  &\textcolor{olive}{\checkmark}\\
      \textbf{\textcolor{purple}{DUST3R}~\cite{dust3r}}                           & 
      \textcolor{olive}{\checkmark} & \textcolor{red}{\ding{55}}      & \textcolor{red}{\ding{55}}  & \textcolor{olive}{\checkmark} & \textcolor{olive}{\checkmark} & \textcolor{olive}{\checkmark} &\textcolor{olive}{\checkmark}\\
      \textbf{\textcolor{purple}{ViPE}~\cite{vipe}}                           & 
      \textcolor{olive}{\checkmark} & \textcolor{red}{\ding{55}}      & \textcolor{red}{\ding{55}}  & \textcolor{olive}{\checkmark} & \textcolor{olive}{\checkmark} & \textcolor{red}{\ding{55}} &\textcolor{olive}{\checkmark}\\
      \midrule
      \textbf{\name{}}                           & 
      \textcolor{olive}{\checkmark} & \textcolor{green}{Open}     & \textcolor{olive}{\checkmark} & \textcolor{olive}{\checkmark} & \textcolor{olive}{\checkmark} & \textcolor{olive}{\checkmark} &\textcolor{olive}{\checkmark}\\
      \bottomrule
    \end{tabular}
    \caption{Comparison of State-of-the-Art SLAM Methods}
    \label{tab:comparison}
    \begin{tablenotes}
    \item \textbf{Columns} indicate: Online operation, Semantic understanding type (Closed=predefined classes, Agnostic=category-agnostic without semantics, Open=open-vocabulary), Geometric Grounding=the ability to textually access and interact with the 3$\mathrm{D}$ map, Odometry estimation, Map reconstruction, Dynamic scene handling and calibration requirement.
    \end{tablenotes}
  \end{threeparttable}
\end{table}
\section{Related works}
\label{sec:related_works}

Recently, SLAM systems extended from purely geometric to high-level and deep scene understanding. We examine existing methodologies from literature based on their capabilities in localization, mapping, semantic understanding, real-time operation, and dynamic scene handling. We categorize SLAM approaches into: \textcolor{traditional}{Geometric SLAM Methods}, \textcolor{semantic}{Offline Open-Vocabulary Mapping}, \textcolor{neural}{Real-time Open-Vocabulary SLAM} and \textcolor{purple}{Feed-forward SLAM} (Table~\ref{tab:comparison}).

\subsection{Geometric SLAM Systems}

Early SLAM systems prioritized geometric understanding, focusing primarily on accurate pose estimation, but often lacks high-level representation and require offline calibration. ORB-SLAM3 \cite{ORB-SLAM3} represents a powerful baseline in visual-inertial SLAM, offering robust multi-map capabilities and loop closure detection across various sensor configurations. While providing excellent odometry and mapping functionality, it cannot interpret scene semantics or handle dynamic objects. RVWO \cite{rvwo} extends classical approaches by introducing dynamic object handling for wheeled mobile robots, yet still lacks semantic understanding capabilities. Kimera \cite{kimera} integrates instance-level semantics into real-time metric mapping but remains constrained to a closed set of pre-defined categories. Similarly, RGBDS-SLAM \cite{rgbdsslam} fuses RGB-D data with semantic segmentation for enhanced reconstruction, yet cannot generalize to novel object categories. SamSLAM \cite{SamSLAM} represents a significant advancement by employing category-agnostic segmentation for mapping and dynamic object handling, but lacks the capability to associate segments with language descriptions or semantic concepts.

\subsection{Offline Open-Vocabulary Scene Understanding}

The emergence of foundation models has enabled open-vocabulary understanding in 3$\mathrm{D}$ scene reconstruction. 
BBQ \cite{bbq} and ConceptGraphs \cite{concept-graphs} leverage large vision-language models to construct semantically rich 3$\mathrm{D}$ scene graphs with natural language grounding. 
HOV-SG \cite{hovsg} extends this paradigm through hierarchical organization of spatial and semantic relationships. 
Meanwhile, OpenScene \cite{openscene} and OpenMask3D \cite{openmask3d} achieve zero-shot 3$\mathrm{D}$ segmentation by distilling CLIP features into 3$\mathrm{D}$ representations. 
Despite their impressive semantic capabilities, these approaches operate exclusively offline and cannot support real-time applications. 
In addition, they lack integrated odometry estimation and cannot handle dynamic scenes, limiting their practical deployment in robotics applications.

\subsection{Real-time Open-Vocabulary SLAM}

Recent works have attempted to bridge the gap between open-vocabulary understanding and real-time SLAM. 
CLIO \cite{clio} introduces an information-theoretic framework that dynamically clusters 3$\mathrm{D}$ primitives according to task-specific language instructions, enabling targeted scene understanding. 
OVO-SLAM \cite{OVO-SLAM} represents another significant advance, integrating CLIP embeddings with Gaussian splatting for real-time open-vocabulary mapping. 
While these approaches successfully combine open-vocabulary capabilities with odometry estimation and mapping, they lack robustness in dynamic scenes.

\subsection{Feed-Forward SLAM}

Traditional reconstruction pipelines require camera calibration and pose estimation before 3$\mathrm{D}$ reconstruction, while recent feed-forward approaches directly regress 3$\mathrm{D}$ geometry from images without prior camera parameters.
DUSt3R~\cite{dust3r} cast pairwise reconstruction as pointmap regression, relaxing rigid projective camera constraints. MASt3R-SLAM~\cite{mast3r} achieved real-time dense SLAM without fixed camera models. VGGT-SLAM~\cite{vggt} incrementally aligned submaps while explicitly optimizing over the SL(4) manifold to account for the 15-DoF projective ambiguity in uncalibrated monocular reconstruction. ViPE~\cite{vipe} extended this to in-the-wild videos, achieving robust intrinsic/extrinsic estimation and near-metric depth at 3--5 FPS. However, feed-forward SLAM systems struggle with dynamic scenes where moving objects violate static assumptions, and lack high-level semantic features for open-vocabulary interaction and grounding.

Despite significant progress in both semantic understanding and geometric mapping, existing systems do not successfully integrate together real-time operation, open-vocabulary semantics, geometric grounding, robust odometry, accurate mapping, and dynamic scene handling simultaneously. 
This integration gap severely limits the deployment of SLAM systems in complex, changing environments where both spatial awareness and rich semantic understanding are required. \name{} introduces a unified SLAM framework that seamlessly combines all these capabilities and can be used with applications as online semantic localization for autonomous robots to perform general-purpose tasks, harvesting data from ego-centric internet videos for training better navigation policies, or interactive open vocabulary augmented reality applications \cite{aria}.

\section{Methodology}
\begin{figure*}[ht]
    \centering
    \includegraphics[width=\textwidth]{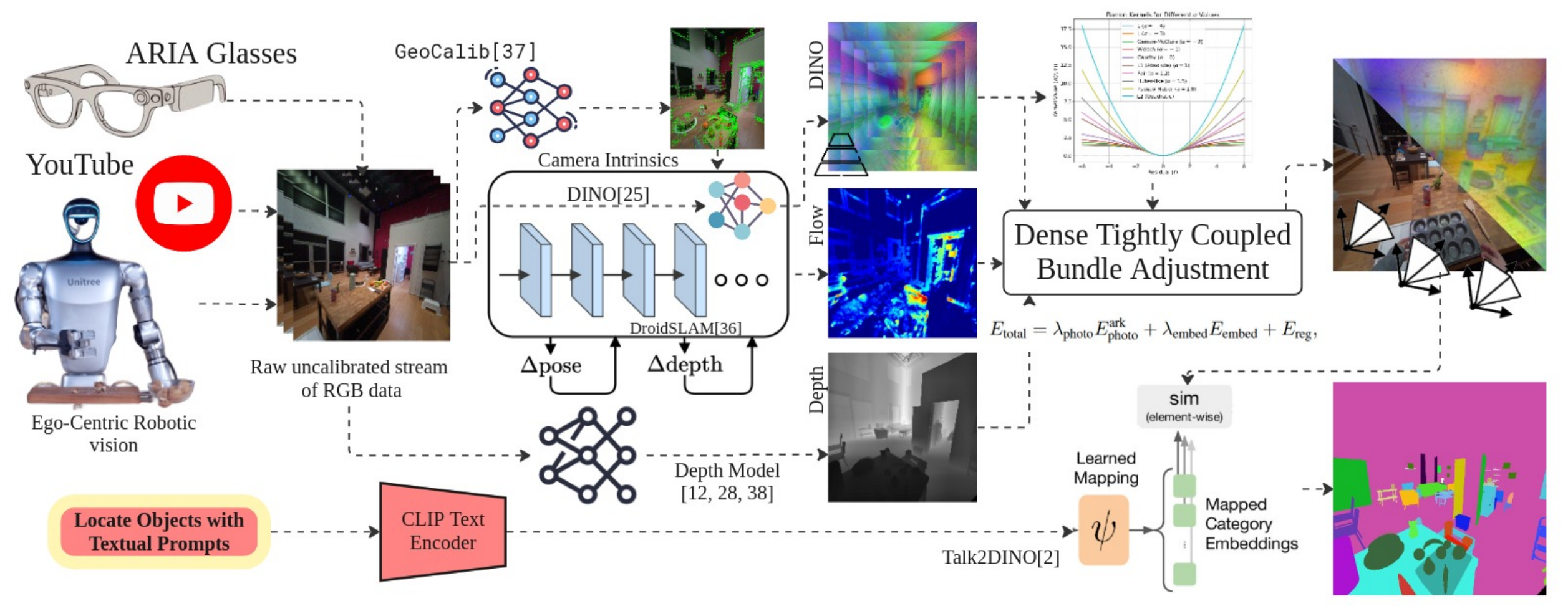} 
    \caption{System pipeline}
    \label{fig:pipeline}
\end{figure*}
We introduce \name{}, a real-time dense visual SLAM system that integrates geometric reconstruction with semantic understanding through foundation model features~\cite{dinov2, dinov3}. Our approach unifies metric depth estimation, camera pose optimization, and dense high-level visual embeddings within a sliding window factor graph optimization framework.
\subsection{System Overview}
\name{} takes as input an uncalibrated monocular RGB video stream and operates at around 8 FPS, producing metrically consistent 3$\mathrm{D}$ reconstructions augmented with open-vocabulary semantic descriptors from DINO features~\cite{dinov2}. Following~\cite{droid, vipe}, the pipeline consists of the following stages:
\begin{enumerate}
    \item  We uniformly sample frames from the input video and estimate focal length and principal point using GeoCalib~\cite{geocalib} without the need of any calibration patterns~\cite{vipe}.
    \item For each incoming frame, we estimate its relative motion to the last keyframe via weighted dense optical flow~\cite{droid, vipe}. Frames exceeding a motion threshold become keyframes and are added to the bundle adjustment graph $\mathcal{G} = (\mathcal{V}, \mathcal{E})$, where $\mathcal{V}$ denotes keyframes and $\mathcal{E}$ denotes pairwise connections~\cite{vipe}.
    \item For each new keyframe, we construct a local sliding window graph in which edges connect frames based on temporal proximity and visual co-visibility. To improve efficiency, all bundle adjustment terms are computed using downsampled images $\tilde{\mathbf{I}} \in \mathbb{R}^{H \times W \times 3}$ with $H = h/8$ and $W = w/8$ for efficiency~\cite{vipe}. For each downsampled keyframe, we compute depth estimates and dense high-level embeddings. Additionally, for every pair of keyframes linked by an edge in the graph, we compute weighted dense optical flow between their downsampled images; these flows are then used to formulate the photometric and motion consistency terms in the optimization.
    \item Dynamic objects can be classified into moving agents and movable objects by agents~\cite{rvwo}. Identifying moving agents is straigt-forward using grounding semantics or object detections, which is also adapted in ~\cite{vipe}. However, detecting movable objects (e.g., moving chairs, cups, books) is challenging as they are identified as static objects. We build on adaptive robust kernels from ~\cite{rvwo, ark} and use the similarity between multi-view invariant DINO features to tune shape parameter $\alpha$ of the Barron function ~\cite{barron} and change the kernel for projection error based on it.
    
    \item The backend performs global bundle adjustment over all keyframes, jointly optimizing camera intrinsics, poses, and scene structure.
    \item The incrementally reconstructed map consists of 3$\mathrm{D}$ of which each has encoded high-level DINO features. At this stage, we decode these embeddings to its original dimension, then by utilizing~\cite{talk2dino} we aligns the textual embeddings of CLIP~\cite{CLIP} to the DINO features through a learned mapping function, which allows to query the 3$\mathrm{D}$ map using textual prompts for open vocabulary segmentation.
    \item  Following~\cite{vipe}, we build local graphs for intermediate (non-key) frames, connecting each to its two nearest keyframes via uni-directional edges. This allows us to estimate their camera poses by aligning them photometrically to nearby keyframes, while avoiding per-frame depth estimation. The process is parallelized across all non-keyframes for efficiency.
    \item Finally, we estimate dense metric depth maps at full resolution $\mathbf{I} \in \mathbb{R}^{h \times w \times 3}$ for all frames, ensuring consistency with optimized poses and intrinsics~\cite{vipe}.
    
\end{enumerate}
\subsection{Dense Visual Feature Extraction}

\subsubsection{Multi-Scale pixel-level DINO Embedding}

For each input frame $\mathbf{I} \in \mathbb{R}^{H \times W \times 3}$, we extract dense pixel-level features using DINO~\cite{dinov2,dinov3}, a self-supervised vision transformer pretrained on large-scale image collections. Unlike methods that operate at a single resolution, we employ a multi-scale pyramid approach to capture features at different levels of detail and provide smoother upsampling from patch level to pixel level embedding.

Given an input image, we generate a pyramid of $S$ scales $\{\sigma_1, \sigma_2, \ldots, \sigma_S\}$ where each scale produces a resized version of the image. For scale $\sigma_s$, we resize the image to $(H_s, W_s)$ where dimensions are aligned to the patch size $p = 14$ of the DINO vision transformer:
\begin{equation}
H_s = \left\lceil \frac{H \cdot \sigma_s}{p} \right\rceil \cdot p, \quad W_s = \left\lceil \frac{W \cdot \sigma_s}{p} \right\rceil \cdot p
\end{equation}

Each scaled image is passed through DINO to extract patch-level visual features. For scale $s$, the output is a feature map:
\begin{equation}
\mathbf{F}_s = \text{DINO}(\mathbf{I}_s) \in \mathbb{R}^{C \times h_s \times w_s}
\end{equation}
where $C$ is the embedding dimension (384 for ViT-S, 768 for ViT-B), and $(h_s, w_s) = (H_s/p, W_s/p)$ is the feature map resolution. By default, we use scales $\{\sigma_s\} = \{2.0, 1.5, 1.0, 0.75\}$ to capture both fine-grained local details and ensure smooth upsampling.

\subsubsection{Pyramid Feature Upsampling}

To obtain dense per-pixel embeddings at the original image resolution, we upsample and blend the multi-scale features. Given the pyramid of feature maps $\{\mathbf{F}_1, \ldots, \mathbf{F}_S\}$, we first upsample each to the target resolution $(H, W)$ using bilinear interpolation:
\begin{equation}
\tilde{\mathbf{F}}_s = \text{Upsample}(\mathbf{F}_s, (H, W)) \in \mathbb{R}^{C \times H \times W}
\end{equation}

We then combine the upsampled features using weighted blending:
\begin{equation}
\mathbf{F} = \frac{\sum_{s=1}^{S} w_s \cdot \tilde{\mathbf{F}}_s}{\sum_{s=1}^{S} w_s}
\end{equation}
where $w_s$ are scale-dependent weights. For weighted blending, we set $w_s = \sigma_s$ to prioritize higher-resolution features. The final dense feature map $\mathbf{F} \in \mathbb{R}^{C \times H \times W}$ provides a smooth view-invariant descriptors for every pixel in the input image.

\subsubsection{PCA-Based Feature Compression}
To reduce memory and computational costs while preserving discriminative information, we apply PCA to compress features from $C$ to $K$ dimensions (we choose $K = 32$).

Given feature vectors $\{\mathbf{f}_i\}_{i=1}^{N}$ sampled from multiple frames, we compute the mean $\boldsymbol{\mu}$ and covariance of these features, then perform SVD on the centered feature matrix $\mathbf{X} = [\mathbf{f}_1 - \boldsymbol{\mu}, \ldots, \mathbf{f}_N - \boldsymbol{\mu}]^\top = \mathbf{U} \boldsymbol{\Sigma} \mathbf{V}^\top$ and retain the top $K$ components $\mathbf{V}_K \in \mathbb{R}^{C \times K}$.
This reduces storage from $C \cdot H \cdot W$ to $K \cdot H \cdot W$ per keyframe while preserving semantic relationships.

\subsection{Depth Estimation and Initialization}

Following~\cite{vipe}, We estimate the metric depth for each keyframe using foundation depth models~\cite{moge,unidepth,metric3dv2}. Given RGB input $\mathbf{I}$ and estimated focal length $f$, the depth model produces a metric depth map $\mathbf{D} \in \mathbb{R}^{H \times W}$. We convert depth to inverse depth (disparity) for numerical stability:
\begin{equation}
d(u, v) = \begin{cases}
\frac{1}{D(u, v)} & \text{if } D(u, v) > 0 \\
0 & \text{otherwise}
\end{cases}
\end{equation}

The initial disparity maps, generated by the depth estimation models, are downsampled by a factor of 8 to match the feature resolution used in bundle adjustment, producing $\mathbf{d} \in \mathbb{R}^{(h/8) \times (w/8)}$. This downsampled representation serves as a geometric prior during optimization. 

\subsection{Joint Bundle Adjustment}

Let $\mathbf{T}_i \in \text{SE}(3)$ denote the world-to-camera transformation for frame $i$, represented as a 6-DOF pose. Camera intrinsics are denoted $\mathbf{K}_q = [f_x, f_y, c_x, c_y]^\top$ for pinhole cameras (extended appropriately for other camera models).

\name{} formulates dense SLAM as a factor graph optimization problem where we jointly optimize camera poses $\{\mathbf{T}_i\}$, dense disparity maps $\{\mathbf{d}_i\}$, and camera intrinsics $\{\mathbf{K}_q\}$. We minimize a weighted sum of photometric, geometric, and semantic consistency terms.

\subsubsection{Dense Photometric Flow Term}

Following DROID-SLAM~\cite{droid}, the core geometric term enforces photometric consistency between frames through dense optical flow constraints. For each edge $(i, j)$ connecting frames at views $(q_i, q_j)$, we project pixels $\mathbf{u} = (u, v)$ from frame $i$ to frame $j$:
\begin{equation}
\mu_{ij} = \Pi_j(\mathbf{T}_j \mathbf{T}_i^{-1} \circ \Pi_i^{-1}(\mathbf{u}, d_i(\mathbf{u})))
\end{equation}
where $\Pi_q$ and $\Pi_q^{-1}$ are the projection and unprojection functions for camera $q$, incorporating both intrinsics $\mathbf{K}_q$.

An optical flow network~\cite{droid} takes the two frames as input and produces a dense flow field $\boldsymbol{\Omega}_{ij} \in \mathbb{R}^{H \times W \times 2}$ at the same resolution as the depth map ($H = h/8$, $W = w/8$). The network constructs a correlation volume with an iterative refinement module and provides a prior flow estimate:
\begin{equation}
\boldsymbol{\Omega}_{ij}^{\text{prior}} = \mu_{ij} - \mathbf{u}
\end{equation}
as initial guidance for the correlation volume. In addition to the flow, a confidence weight map $w[\mathbf{u}]$ is estimated to reflect the reliability of the flow estimation.

The dense flow term measures the reprojection error across all $h \times w$ pixels in the depth map $\mathbf{D}_i$:
\begin{equation}
E_{\text{photo}}(\mathbf{T}_i, \mathbf{T}_j, \mathbf{D}_i; K_q) = \sum_{\mathbf{u}} w(\mathbf{u}) \cdot \left\| \boldsymbol{\Omega}_{ij}^{\text{prior}} - \boldsymbol{\Omega}_{ij}(\mathbf{u}) \right\|^2
\end{equation}
where $\boldsymbol{\Omega}_{ij}(\mathbf{u})$ are the target flow vectors from the optical flow network, $w(\mathbf{u})$ are per-pixel confidence weights from the optical flow network, and $K$ denotes the used camera model parameters. This formulation directly compares the geometric reprojection from poses and depth against the learned dense correspondences, enabling robust optimization even in challenging scenarios with limited texture.

\subsubsection{Embedding Similarity Term}

To incorporate semantic consistency, we introduce a novel embedding similarity term that enforces feature alignment across views while respecting geometric constraints. This term enables \name{} to leverage high-level deep visual information from foundation models during optimization.

For an edge $(i, j)$ with source view $q_i$ and target view $q_j$, let $\mathbf{Z}_i \in \mathbb{R}^{K \times H \times W}$ and $\mathbf{Z}_j \in \mathbb{R}^{K \times H \times W}$ denote the dense embedding maps (PCA-compressed). We project each pixel from the source frame to the target frame using the current depth and pose estimates, and then measure the cosine similarity between the source embedding and the bilinearly sampled target embedding.

\textbf{Bilinear Sampling} Given continuous coordinates $\mathbf{u} = (u, v)$ in the target frame $j$, we sample the target embedding using bilinear interpolation from the four neighboring pixels:
\begin{equation}
\mathbf{z}_j(\mathbf{u}) = \sum_{(l, m) \in \mathcal{N}(\mathbf{u})} w_{lm}(\mathbf{u}) \cdot \mathbf{Z}^{j}_{lm}
\end{equation}
where $\mathcal{N}(\mathbf{u}) = \{(u_0, v_0), (u_0+1, v_0), (u_0, v_0+1), (u_0+1, v_0+1)\}$ with $u_0 = \lfloor u \rfloor$ and $v_0 = \lfloor v \rfloor$ denotes the four neighboring pixel locations, $\mathbf{Z}^{j}_{lm} \in \mathbb{R}^K$ are the PCA-compressed DINO features at pixel $(l, m)$ in frame $j$, and $w_{lm}(\mathbf{u})$ are the bilinear weights. Using relative indices where $(0,0)$ corresponds to $(u_0, v_0)$, the weights are:
\begin{align}
w_{00} &= \alpha\beta, \quad w_{01} = \alpha(1 - \beta) \\
w_{10} &= (1 - \alpha)\beta, \quad w_{11} = (1 - \alpha)(1 - \beta)
\end{align}
with $\alpha = u - u_0$ and $\beta = v - v_0$ being the fractional offsets within the pixel cell.

 We normalize embeddings after sampling to ensure invariance to feature magnitude. Let $\mathbf{s} = \mathbf{Z}_i / \|\mathbf{Z}_i\|$ be the normalized source embedding in view $i$ (directly indexed, no sampling needed) and $\mathbf{t} = \mathbf{z}_j / \|\mathbf{z}_j\|$ be the normalized sampled target embedding in view $j$. The cosine similarity is:
\begin{equation}
cs_{ij} = \frac{\mathbf{Z}_i^\top \mathbf{z}_j}{\|\mathbf{Z}_i\| \cdot \|\mathbf{z}_j\|}
\end{equation}

We define the embedding residual using one of two formulations:

\textit{Angular residual}:
\begin{equation}
r_{\text{embed}}(\mathbf{u}_i) = 1 - cs_{ij}
\end{equation}

\textit{Photometric-style residual}:
\begin{equation}
r_{\text{embed}}(\mathbf{u}_i) = \lambda_{\text{embed}} \sqrt{2(1 - cs_{ij})}
\end{equation}
where $\lambda_{\text{embed}}$ is a scale factor (in our experiments, $\lambda_{\text{embed}}=2$). The photometric-style formulation produces residuals that behave similarly to intensity differences in traditional photometric SLAM.

The full embedding similarity term is:
\begin{equation}
E_{\text{embed}}(\mathbf{T}_i, \mathbf{T}_j, \mathbf{d}_i) = \sum_{\mathbf{u}} w_{\text{embed}}(\mathbf{u}) \cdot r_{\text{embed}}^2(\mathbf{u})
\end{equation}
where $w_{\text{embed}}(\mathbf{u})$ incorporates both the photometric flow confidence and embedding validity masks.

\textbf{Jacobian with Respect to Pose and Depth.} The chain rule gives:
\begin{equation}
\frac{\partial r_{\text{embed}}}{\partial \boldsymbol{\xi}} = \frac{\partial r_{\text{embed}}}{\partial cs} \cdot \frac{\partial cs}{\partial \mathbf{z}_j} \cdot \frac{\partial \mathbf{z}_j}{\partial \mathbf{u}_{ij}} \cdot \frac{\partial \mathbf{u}_{ij}}{\partial \boldsymbol{\xi}}
\end{equation}
where $\boldsymbol{\xi}$ represents pose or depth parameters and $\frac{\partial \mathbf{z}_j}{\partial \mathbf{u}_{ij}}$ represents the derivative of the embeddings with respect to the pixel location according to the bilinear sampling above.

For the angular residual, $\frac{\partial r_{\text{embed}}}{\partial cs} = -1$. While for the photometric residual:
\begin{equation}
\frac{\partial r_{\text{embed}}}{\partial c} = -\frac{\lambda_{\text{embed}}^2}{r_{\text{embed}} + \epsilon}
\end{equation}

The derivative of cosine similarity with respect to the raw sampled feature (before normalization) is:
\begin{equation}
\frac{\partial cs}{\partial \mathbf{z}_j} = \frac{\mathbf{s} - cs \cdot \mathbf{t}}{\|\mathbf{z}_j\|}
\end{equation}
This accounts for the normalization applied after sampling.

\subsubsection{Adaptive Robust Kernel for Dynamic Regions}

Real-world ego-centric videos contain both static backgrounds and dynamic objects such as humans, pets, and manipulable objects. While the dense flow term $E_{\text{photo}}$ already down-weights unreliable regions through confidence maps, it assumes a fixed noise distribution over all residuals. To further suppress the influence of outliers induced by dynamic regions and subtle non-rigid motion, we adopt an adaptive robust kernel based on the Barron loss~\cite{barron}, whose shape is modulated by multi-view DINO feature similarity~\cite{rvwo, ark}.

Let $cs_{ij}(\mathbf{u})$ denote the cosine similarity between the DINO embeddings of views $i$ and $j$ at location $\mathbf{u}$, as defined in the embedding similarity term. Pixels on static surfaces yield high $cs_{ij}(\mathbf{u})$, whereas pixels on movable or independently moving objects produce low similarity due to appearance changes and occlusions.

We map the similarity into a \emph{dynamic score} $\alpha_{ij}(\mathbf{u}) \in [\alpha_{\text{dynamic}}, \alpha_{\text{static}}]$ via a sigmoid function:
\begin{equation}
\alpha_{ij}(\mathbf{u}) = \frac{\alpha_{\text{dynamic}} - \alpha_{\text{static}}}{1+\exp(\frac{cs_{ij}(\mathbf{u})-\kappa}{\tau})} + \alpha_{\text{static}} 
\end{equation}
where $\tau$ controls the sharpness of the transition, and $\kappa$ is a similarity threshold. This dynamic score will be used to adjust the kernel of the error term.

\begin{figure}[t]
    \centering
    \includegraphics[width=0.45\textwidth]{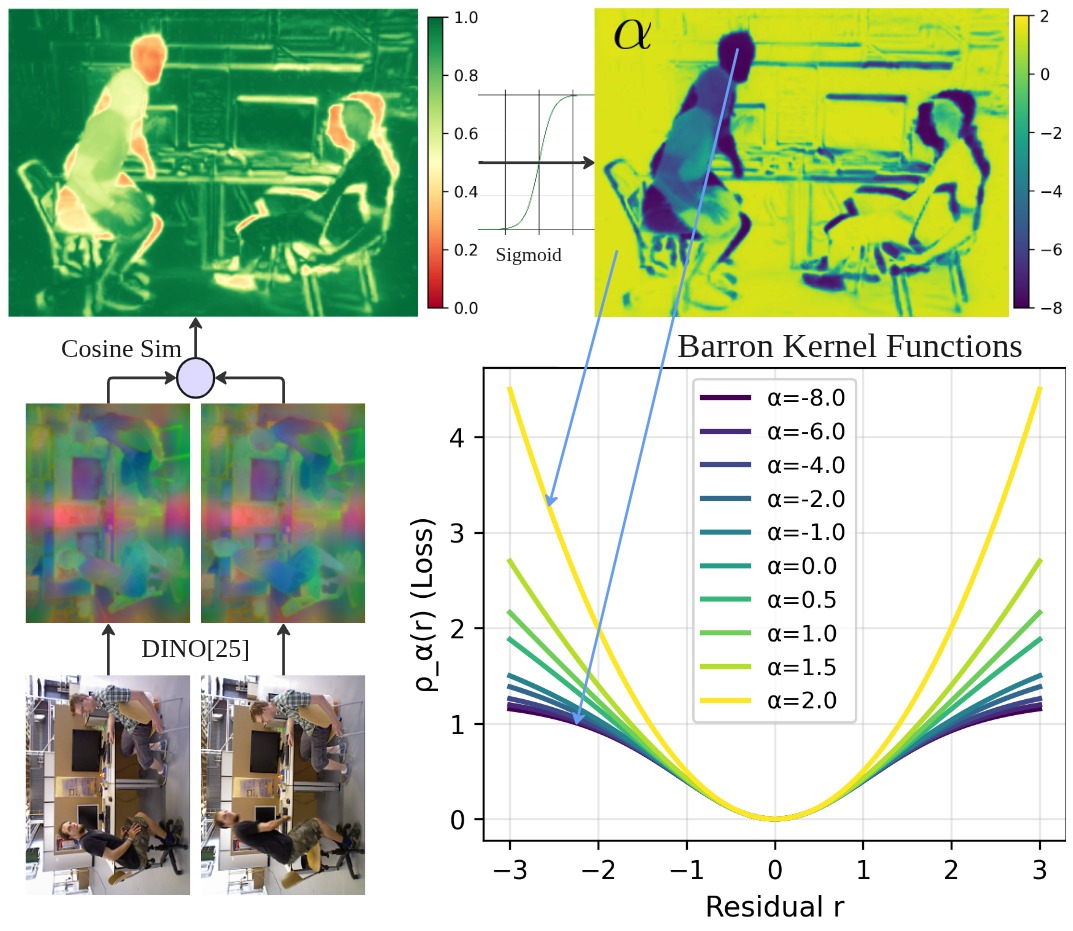}
    \caption{Adaptive robust kernels based on Barron's function~\cite{barron} in Bundle Adjustment. Shape parameter $\alpha$ is determined by the cosine similarity between multiview high-level visual features~\cite{dinov2}}
    \label{fig:barron}
\end{figure}

Following~\cite{barron}, we use the general robust loss $\rho_\alpha(r)$ with shape parameter $\alpha$ and scale $c$:
\begin{equation}
\rho_\alpha(r) =
\frac{|\alpha - 2|}{\alpha}
\left[
\left(
\frac{(r / c)^2}{|\alpha - 2|} + 1
\right)^{\alpha/2}
- 1
\right],
\label{eq:barron-loss}
\end{equation}
which recovers several standard losses for particular $\alpha$ (e.g., $\alpha = 2$ for $\ell_2$, $\alpha = 1$ for Huber-like, and $\alpha \rightarrow 0$ for Cauchy). Instead of using a fixed $\alpha$, we use the dynamic score computed earlier $\alpha_{ij}(\mathbf{u})$ resulting in an adaptive kernel based on the embedding similarity between the considered pixels (See Fig. ~\ref{fig:barron}).

with $\alpha_{\text{static}} \approx 2$ and $\alpha_{\text{dynamic}} \leq 0$.
The corresponding adaptive robust kernel for the photometric-flow term is
\begin{equation}
E_{\text{photo}}^{\text{ark}} =
\sum_{(i,j) \in \mathcal{E}} \sum_{\mathbf{u}}
w(\mathbf{u}) \,
\rho_{\alpha_{ij}(\mathbf{u})}
\big(E_{\text{photo}}(\mathbf{u}_{ij})\big),
\end{equation}

In practice, we implement the robust kernel within Gauss--Newton using an iteratively reweighted least squares (IRLS) formulation. The influence function of $\rho_\alpha$ and the corresponding weight for residual $r=E_{\text{photo}}$ are
\begin{equation}
w_{\text{ark}}(r;\alpha) = \frac{\psi_\alpha(r)}{\max(r, \varepsilon)} \quad ;\quad\psi_\alpha(r) = \frac{\partial \rho_\alpha(r)}{\partial r},
\end{equation}
with a small constant $\varepsilon$ to avoid division by zero. For each pixel $\mathbf{u}$ we obtain $w_{\text{ark}}(\mathbf{u}) = w_{\text{ark}}(r(\mathbf{u}); \alpha_{ij}(\mathbf{u}))$ and fold it into the overall per-pixel weight by
\begin{equation}
\tilde{w}(\mathbf{u}) = w(\mathbf{u}) \cdot w_{\text{ark}}(\mathbf{u})
\end{equation}
The adaptive robust kernel therefore down-weights residuals both in low-confidence regions (via $w$) and in semantically dynamic regions (via $\alpha_{ij}(\mathbf{u})$), while preserving high influence for static pixels.
\subsubsection{Disparity Regularization Term}

Following~\cite{vipe}, we add a soft regularization term to stabilize depth estimates and incorporate prior knowledge from foundation depth models:
\begin{equation}
E_{\text{reg}}(\mathbf{d}_i) = \alpha_{\text{disp}} \sum_{\mathbf{u}} \| \mathbf{d}_i(\mathbf{u}) - \mathbf{d}_i^{\text{prior}}(\mathbf{u}) \|^2
\end{equation}
where $\mathbf{d}_i^{\text{prior}}$ is the disparity from the foundation depth model and $\alpha_{\text{disp}} = 1.0$ is the regularization weight. This term prevents drift while allowing the optimization to refine depth estimates based on multi-view consistency.

\subsubsection{Total Objective and Optimization}



\begin{equation}
E_{\text{total}} =
\lambda_{\text{photo}} E_{\text{photo}}^{\text{ark}}
+ \lambda_{\text{embed}} E_{\text{embed}}
+ E_{\text{reg}},
\end{equation}
which we minimize with Gauss--Newton over camera poses $\{\mathbf{T}_i\}$, disparities $\{\mathbf{d}_i\}$, and intrinsics $\{\mathbf{K}_q\}$. The high-level optimization loop is summarized in Alg.~\ref{alg:ark-ba}.

\begin{algorithm}[t]
    \caption{Joint Bundle Adjustment with ARK}
    \label{alg:ark-ba}
    \begin{enumerate}
        \item \textbf{Inputs \& Initialization:} keyframes $\mathcal{V}$, edges $\mathcal{E}$, initial poses $\{\mathbf{T}_i\}$,
        intrinsics $\{\mathbf{K}_q\}$, disparities $\{\mathbf{d}_i\}$, depth priors $\{\mathbf{d}_i^{\text{prior}}\}$,
        and DINO embeddings $\{\mathbf{Z}_i\}$.

        \item \textbf{Repeat until convergence or max iterations:}
        \begin{enumerate}
            \item Build normal equations $\mathbf{H}, \mathbf{b}$ by looping over all edges $(i,j) \in \mathcal{E}$ and pixels:
            project pixels, compute flow and embedding residuals, derive DINO-based similarity $cs_{ij}$,
            obtain adaptive shape $\alpha_{ij}$ and robust weight $w_{\text{ark}}$, and accumulate their
            contributions into the cost function $E_{\text{Total}}$.
            \item Add disparity regularization $E_{\text{reg}}$ to $\mathbf{H}, \mathbf{b}$, solve the
        Gaussian-Newton system, and update $\{\mathbf{T}_i\}$, $\{\mathbf{K}_q\}$, $\{\mathbf{d}_i\}$ via the respective retractions.
        \end{enumerate}

        \item \textbf{Output:} optimized poses, intrinsics, and disparities
        $\{\mathbf{T}_i\}$, $\{\mathbf{K}_q\}$, $\{\mathbf{d}_i\}$.
    \end{enumerate}
\end{algorithm}

\section{Experiments}
\label{sec:experiments}



We evaluate our system across these key capabilities: SLAM accuracy, semantic understanding, and dynamic environment handling. 
Experiments are conducted on standard datasets including Replica \cite{replica}, TUM-RGBD \cite{tum-rgbd}, and ARIA \cite{aria}, with comparisons against state-of-the-art methods. We denote the variant of our pipeline that incorporates only the embedding error term as $\text{KM-ViPE}$, and the full pipeline that includes both the embedding error term and the adaptive kernel tuning as $\text{KM-ViPE}_{ark}$. 
\begin{table}[htbp]
    \small
    \setlength{\tabcolsep}{1.5pt} 
    \caption{SLAM Performance Comparison on TUM-RGBD Dataset in cm (ATE)}
    \begin{tabular}{lccccccccc}
        \toprule
        \textbf{Method} 
        & \rotatebox{90}{fr3/w/xyz} 
        & \rotatebox{90}{fr3/w/rpy} 
        & \rotatebox{90}{fr3/w/hs} 
        & \rotatebox{90}{fr3/w/static}
        & \rotatebox{90}{fr3/s/xyz}
        & \rotatebox{90}{fr3/s/rpy}
        & \rotatebox{90}{fr3/s/hs}
        & \rotatebox{90}{fr3/s/static}
        & \rotatebox{90}{average} \\
        \midrule
        Dyna-SLAM~\cite{dynaslam2} & 1.64 & 3.54 & 2.96 & 0.68 & 1.27 & - & 1.86 & - & 2.0\\
        DLD-SLAM~\cite{dldslam}  & 1.85 & 4.24 & 2.19 & 0.56 & - & - & - & - & 2.21\\
        V3D-SLAM~\cite{v3d} & \textbf{1.53} & 7.81 & \textbf{2.29} & 0.65 & 0.87 & \textbf{1.69} & \textbf{1.47} & 0.58 & 2.1 \\
        DGS-SLAM~\cite{dgs}  & 4.1 & - & 5.5 & 0.6  & - & - & 4.4 & - & 3.65\\
        RoDyn-SLAM~\cite{rodyn} & 8.3 & - & 5.6 & 1.7 & - & - & 2.7 & - & 4.58\\
        ViPE (SAM)~\cite{vipe}  & 2.35 & 6.27 & 10.83 & \textbf{0.51} & 5.41 & 3.3 & 3.53 &  0.53 & 4.1 \\
        $\text{KM-ViPE}$  & 1.9 & \textbf{3.50} & 3.1 & 0.55 & \textbf{1.15} & 2.72 & 1.6 &  0.53 & 1.9  \\
        $\text{KM-ViPE}_{ark}$ & 1.82 & 3.52 & 2.9 & 0.54 & \textbf{1.15} & 2.71 & 1.58 &  \textbf{0.51} & \textbf{1.84}  \\
        \bottomrule
    \end{tabular}
    \label{tab:slam-tum}
\end{table}

\subsection{SLAM Performance}
Our embedding error term and adaptive robust kernel were designed to improve SLAM performance in dynamic environments, where the motion of objects or agents often leads to incorrect data associations. These errors are either compensated for by the embedding term or attenuated during optimization through adaptive kernel tuning. To evaluate the effectiveness of our approach, we conducted experiments on the dynamic sequences of the TUM RGB-D~\cite{tum-rgbd} dataset. The results demonstrate that our pipeline achieves state-of-the-art performance, outperforming existing dynamic SLAM methods in the Absolute Trajectory Error (ATE). While ViPE~\cite{vipe} employs foundation models (e.g., Grounding DINO~\cite{gdino} and SAM~\cite{sam}) for dynamic object masking and depends on manual specification of dynamic object classes, our pipeline attains better performance despite requiring substantially fewer resources. Our method is fully autonomous and uses only DINO embeddings obtained directly from the frame.

To ensure that our pipeline does not underperform relative to ViPE on static sequences, we evaluated both methods on the Replica~\cite{replica} dataset. The results show that our pipeline achieves superior average performance across the static sequences.

\begin{table}[t]
    \centering
    \setlength{\tabcolsep}{1.2pt}
    \renewcommand{\arraystretch}{0.9}
    \caption{SLAM Performance Comparison on Replica Dataset in cm (ATE)}
    \label{tab:slam-replica}
    \resizebox{\columnwidth}{!}{%
        \begin{tabular}{lcccccccc|c}
            \toprule
            & \multicolumn{8}{c}{Replica}\\
            \cmidrule(lr){2-9}
            \textbf{Method} 
            & \rotatebox{90}{Room 0} 
            & \rotatebox{90}{Room 1} 
            & \rotatebox{90}{Room 2} 
            & \rotatebox{90}{Office 0} 
            & \rotatebox{90}{Office 1} 
            & \rotatebox{90}{Office 2} 
            & \rotatebox{90}{Office 3} 
            & \rotatebox{90}{Office 4} 
            & \rotatebox{90}{Average}  \\
            \midrule
            VIPE~\cite{vipe}    & \textbf{2.40} & 3.75 & \textbf{3.92} & 6.37 & 14.49 & \textbf{6.05} & \textbf{1.89} & \textbf{1.88} & 5.09 \\
            KM-ViPE             & 3.80    & \textbf{1.66}    & 5.78    & 4.13    & 11.18    & 8.81    & 3.13    & 3.28    & 5.22    \\
            KM-ViPE$_{ark}$       & 3.25    & \textbf{1.66}    & 5.16    & \textbf{4.03}    & \textbf{10.30}    & 8.30    & 3.13    & 3.26   & \textbf{4.88}    \\
            \bottomrule
        \end{tabular}
    }%
\end{table}

\subsection{3D Semantic Segmentation}
One of the features of our approach is the ability to build a semantically-aware map during real-time operation. To assess this capability, the following section provides an evaluation of semantic segmentation performance.

It worth mentioning that to enable training from internet-scale data our method operates under a very different and more challenging setting compared to others open-vocabulary semantic mapping approaches \cite{openscene, concept-graphs, bbq, Open-Nerf, hovsg, Open3dis, OVO-SLAM}, i.e. it takes only a monocular RGB video as input, without requiring known camera parameters, camera poses, a reconstructed point cloud, or depth information. 

In contrast, OpenScene~\cite{openscene}, ConceptGraphs~\cite{concept-graphs}, BBQ~\cite{bbq}, HOV-SG~\cite{hovsg}, OpenNeRF~\cite{Open-Nerf}, and Open3DIS~\cite{Open3dis} all require known camera poses, camera parameters, and depth data (with OpenScene and Open3DIS also requiring a pre-computed point cloud), while OVO-SLAM~\cite{OVO-SLAM}, still depends on known camera parameters and depth to function.  

Furthermore, with the exception of OVO-SLAM, none of these prior works are designed for real-time operation. To the best of our knowledge, no existing pipeline can produce a dense 3D semantic map from a single video in real-time without camera parameters, poses, or depth.

Tables~\ref{tab:semantics-replica}-\ref{tab:semantics-aria} contain semantic mapping results for our method\footnote{We put it here for reference and warn a reader to make direct comparison of obtained results with numbers reported for others open-vocabulary semantic mapping approaches also because of the reason mentioned above.}, evaluated using the OSMA-bench~\cite{osma-bench} protocol. We query the CLIP~\cite{CLIP} model with the template \emph{``There is {label} in the scene''} to obtain text embeddings, which are then projected into the DINOv2~\cite{dinov2} latent space using Talk2DINO~\cite{talk2dino}. We obtain predicted labels for each point by computing the cosine similarity between all point embeddings and all label phrase embeddings, followed by an argmax operator. Finally, we match our estimated point cloud to the vertices of the ground-truth meshes using a KD-tree search with 5 neighbors. 

Experiments have been conducted on eight scenes from the Replica dataset~\cite{replica} \texttt{office0-4, room0-2}. These scenes contain 51 classes, which are divided into three equal groups {head, common, tail} based on the number of points in each class, following the methodology proposed by OpenNeRF~\cite{Open-Nerf}. We use the semantic point clouds from the Replica dataset as ground truth.
\begin{figure}[ht]
    \centering
    \includegraphics[width=0.45\textwidth]{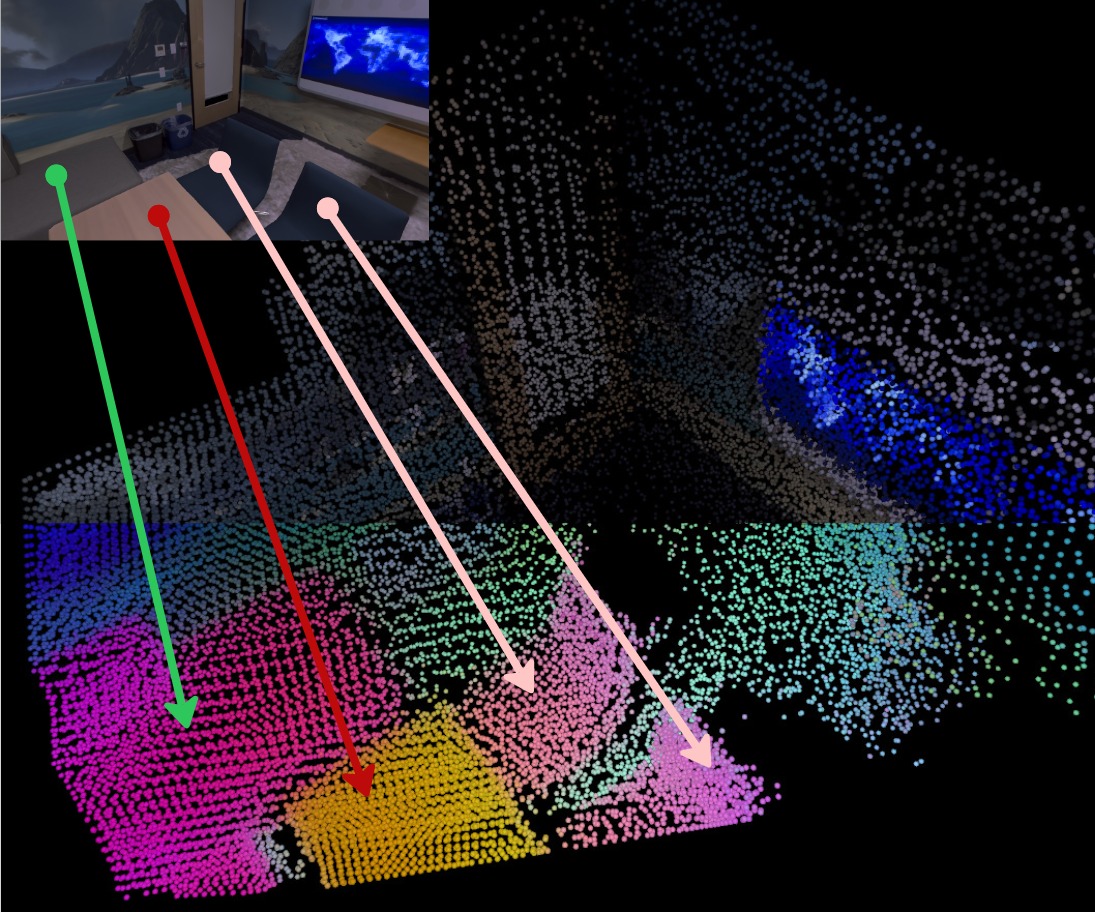}
    \caption{3D Replica Point Cloud with aligned fused high level embeddings, visualized using PCA colorization}
    \label{fig:vis_embed_pcd}
\end{figure}

\begin{table}[h]
    \centering
    \caption{Semantic Segmentation Evaluation on Replica}
    \small
    \begin{tabular}{lccc}
        \toprule
        Scene & mIoU & fmIoU & mAcc \\
        \midrule
        room0 & 6.9 & 11.6 & 14.9 \\
        room1 & 5.9 & 10.9 & 18.6 \\
        room2 & 6.8 & 7.5 & 17.0 \\
        office0 & 9.6 & 16.2 & 19.5 \\
        office1 & 3.0 & 2.7 & 12.6 \\
        office2 & 2.2 & 3.9 & 4.7 \\
        office3 & 4.8 & 11.0 & 16.3 \\
        office4 & 3.7 & 8.2 & 5.3 \\
        \hline
        head & 7.1 & 8.2 & 18.4 \\
        common & 3.5 & 3.7 & 6.4 \\
        tail & 0.9 & 1.0 & 7.9 \\
        \midrule
        overall & 3.8 & 7.9 & 10.9 \\
        \bottomrule
    \end{tabular}
    \label{tab:semantics-replica}
\end{table}


To further validate our approach, we perform evaluations on three dynamic sequences from the Aria Digital Twin\cite{aria} dataset: \small{\texttt{Apartment\_clean\_seq134\_M1292, Apartment\_meal\_skeleton\_seq133\_M1292, Apartment\_multiskeleton\_party\_seq102\_M1292}}. For these sequences, we reconstruct the ground truth point cloud from the image data, explicitly removing dynamic objects to create a static representation for evaluation. This dataset comprises 103 semantic classes, which we similarly divide into head, common, and tail groups based on point frequency, ensuring a consistent evaluation protocol across both datasets.

\begin{table}[ht]
    \centering
    \small
    \caption{Semantic Segmentation Evaluation on Aria}
    \begin{tabular}{lccc}
        \toprule
        Scene & mIoU & fmIoU & mAcc \\
        \midrule
        seq102  & 1.9 & 5.8 & 4.6 \\
        seq134  & 1.8 & 7.0 & 4.7 \\
        seq133  & 1.1 & 5.0 & 4.1 \\
        \hline
        head & 5.1 & 7.1 & 12.5 \\
        common & 0.1 & 0.1 & 0.8 \\
        tail & 0.0 & 0.0 & 0.2 \\
        \midrule
        overall & 1.7 & 6.5 & 4.4 \\
        \bottomrule
    \end{tabular}
    \label{tab:semantics-aria}
\end{table}

\section{Conclusion}
We introduced KM-ViPE, a real-time open-vocabulary semantic SLAM system designed for robust functioning in dynamic environments by integrating geometry with deep visual semantic features for improved data associations, benefiting from training using raw internet-scale video recording from monocular RGB cameras with no extra information on camera parameters or poses. By unifying vision, language, and geometry, KM-ViPE moves toward practical spatial intelligence for embodied agents, enabling navigation and ego-centric perception in demand for autonomous robotic systems and AR/VR solutions.

\section{Future Work}

A promising direction for future work lies in enhancing the quality of semantic segmentation. While our current method provides a foundational framework that already demonstrates competitive results, there is a significant potential for further improvement by incorporating more advanced 2D vision foundation models, fine-tuning strategies specifically for dense prediction tasks or benefiting from very recent advancements in encoder models designed for dynamic environments \cite{Araslanov:2025:FlowFeat}. 

\bibliographystyle{IEEEtran}
\bibliography{IEEEabrv,IEEE_resources}



\end{document}